\documentclass{article} 
\usepackage{iclr2025_conference,times}

\usepackage{amsmath,amsfonts,bm}

\def\eqref#1{equation~\ref{#1}}

\def\1{\bm{1}}

\DeclareMathAlphabet{\mathsfit}{\encodingdefault}{\sfdefault}{m}{sl}
\SetMathAlphabet{\mathsfit}{bold}{\encodingdefault}{\sfdefault}{bx}{n}

\usepackage{hyperref}
\hypersetup{
    colorlinks=true,
    linkcolor=red,
    citecolor=green,
    urlcolor=red
}
\usepackage{url}
\usepackage{graphicx}
\usepackage{xcolor}
\usepackage{algorithm}
\usepackage{algpseudocode}
\usepackage{wrapfig}

\title{MIDAS: \underline{M}ultimodal \underline{I}nteractive \underline{D}igital-hum\underline{A}n \underline{S}ynthesis via Real-time Autoregressive Video Generation}

\author{Ming Chen$^{1}$\thanks{Equal Contribution.} \qquad
    Liyuan Cui$^{1,2}$\footnotemark[1] \qquad 
    Wenyuan Zhang$^{1,3}$\footnotemark[1] \qquad
    Haoxian Zhang$^{1}$
     \\ 
    \textbf{
    Yan Zhou$^{1}$ \qquad
    Xiaohan Li$^{1}$ \qquad
    Songlin Tang$^{1}$ \qquad
    Jiwen Liu$^{1}$ \qquad 
    } \\
    \textbf{
    Borui Liao$^{1}$ \qquad
    Hejia Chen$^{1}$ \qquad
    Xiaoqiang Liu$^{1}$ \qquad
    Pengfei Wan$^{1}$
    }
\\
{}
\\
$^{1}$Kling Team, Kuaishou Technology \quad $^{2}$Zhejiang University \quad $^{3}$ Tsinghua University
}

\iclrfinalcopy 
\begin{document}

\maketitle

\begin{abstract}

Recently, interactive digital human video generation has attracted widespread attention and achieved remarkable progress. However, building such a practical system that can interact with diverse input signals in real time remains challenging to existing methods, which often struggle with heavy computational cost and limited controllability. 
In this work, we introduce an autoregressive video generation framework that enables interactive multimodal control and low-latency extrapolation in a streaming manner. With minimal modifications to a standard large language model (LLM), our framework accepts multimodal condition encodings including audio, pose, and text, and outputs spatially and semantically coherent representations to guide the denoising process of a diffusion head. To support this, we construct a large-scale dialogue dataset of approximately 20,000 hours from multiple sources, providing rich conversational scenarios for training. We further introduce a deep compression autoencoder with up to 64$\times$ reduction ratio, which effectively alleviates the long-horizon inference burden of the autoregressive model.
Extensive experiments on duplex conversation, multilingual human synthesis, and interactive world model highlight the advantages of our approach in low latency, high efficiency, and fine-grained multimodal controllability. 
Project Page: \url{https://chenmingthu.github.io/milm/}.

\end{abstract}

\section{Introduction}
\label{sec:Introduction}
Condition-driven human video generation transforms static portraits into dynamic, interactive virtual avatars that synchronize speech with natural facial expressions, body movements, and emotional cues~\citep{qi2025chatanyone,lin2025omnihuman,xu2024vasa}. Such technologies enhance digital communication by making human-AI interactions more engaging and natural, and opens promising avenues for future applications such as virtual education and creative media. To this end, a practical system must simultaneously satisfy three demands: accept and respond to a broad spectrum of input signals, enable low-latency interaction, and preserve visual and identity consistency over long generation horizons. These requirements pose substantial challenges to existing approaches. 


To meet these requirements, extensive efforts have been devoted across multiple directions. GAN-based approaches~\citep{guo2024liveportrait, qi2025chatanyone} produce high-fidelity local facial details but struggle with global expressiveness and smooth temporal transitions. Video diffusion transformers~\citep{tian2024emo, tian2025emo2, cui2024hallo3, wei2025mocha} achieve high-quality generation but rely on bidirectional attention and iterative denoising, which makes them unsuitable for real-time applications. Although various techniques have been proposed to accelerate diffusion inference~\citep{lu2025dpm,zhai2024motion}, they still face trade-offs among generality, fidelity, and speed. Moreover, most existing methods are limited to unimodal control signals such as audio or text, and require all control parameters to be specified before generation begins, which significantly limits the richness and interactivity of digital human video synthesis.

Recently, autoregressive models offer a promising solution to these limitations by generating tokens conditioned on historical inputs~\citep{deng2025nova,teng2025magi}. This paradigm not only enables flexible interaction by adjusting future predictions based on past frames, but also supports arbitrary-length video synthesis while avoiding the computational overhead of bidirectional attention during inference. In parallel, recent advances in multimodal large language models (MLLMs)~\citep{liu2023llava,wu2024nextgpt} show remarkable potential in bridging generation and understanding, driven by conditioning transformers on diverse multimodal inputs. These developments motivate us to unify language modeling with generative video frameworks into an end-to-end multimodal autoregressive architecture for digital human video generation.


In this work, we introduce \textbf{MIDAS}, a \textbf{m}ultimodal \textbf{i}nteractive \textbf{d}igital hum\textbf{a}n \textbf{s}ynthesis framework that supports low-latency inference, multimodal conditioning, and open-ended generation. The core of our framework is an autoregressive large language model (LLM) that predicts the evolution of video frames in latent space guided by multimodal inputs. Specifically, we design a multimodal condition projector that encodes diverse input signals, including audio, pose, and text, into a shared latent space. These multimodal tokens are concatenated with frame tokens to form the input sequence for the autoregressive model. The frame tokens are encoded via a carefully designed Deep Compression Autoencoder (DC-AE) with a 64$\times$ spatial reduction ratio, which significantly reduces the extrapolation burden of the autoregressive backbone. The autoregressive model outputs hidden states that capture both spatial and semantic coherence, which are then passed to an external diffusion head to render high-quality video frames. During training, the autoregressive model and diffusion head are jointly optimized in a teacher forcing manner, while the inference proceeds via next-frame prediction, enabling real-time and streamable animation generation. To further mitigate exposure bias, we introduce a controlled noise injection mechanism that corrupts context frames with Gaussian noise of varying magnitudes, thereby simulating degraded predictions and teaching the model to recover from imperfect histories during inference. We additionally construct a large-scale dialogue dataset of approximately 20,000 hours, collected from both online sources and custom-recorded sessions. These dialogues are carefully segmented, annotated and post-processed, providing a strong foundation for our model in general digital human dialogue scenarios. We validate the effectiveness of our method through wide experiments on duplex conversation and multi-language human synthesis. We also demonstrate its potential to incorporate richer control signals to act as a general interactive world model. In summary, we make the following contributions:


\begin{itemize}
    \item \textbf{Multimodal Control via Instruction Tokens.} We construct a multimodal architecture that encodes diverse control signals (text, audio, and pose) into global instruction tokens. These tokens condition the autoregressive model for video generation, enabling coherent and expressive latent frame progression over time.
    \item \textbf{Causal Latent Prediction with Diffusion Rendering.} Leveraging the causal structure of autoregressive frameworks, our system predicts one frame's latent tokens at a time. Each predicted representation is processed by a lightweight diffusion head, which recovers high-quality frames with only a few denoising steps. This design enables flexible-length, low-latency video generation ideal for interactive applications.
    \item \textbf{Efficient Representation with High Compression Autoencoder.} We introduce an deep compression autoencoder with up to 64$\times$ reduction ratio. This compact latent representation significantly reduces the number of video tokens, which lowers the computational demands on the autoregressive model while retaining reconstruction fidelity.
\end{itemize}

\section{Related Work}
\label{sec:Related}
\textbf{Speech-driven Talking Face Generation.} 
Animating human portraits represents a challenging frontier in computer vision. Traditional methods~\citep{zhou2020makelttalk, thies2020neural, suwajanakorn2017synthesizing} rely on intermediate representations such as 3D face models or 2D landmarks, which require complex preprocessing pipelines and restrict the generalization of identity. Early approaches~\citep{guo2024liveportrait, qi2025chatanyone} mainly focuses on manipulating fine-grained facial details (gaze, lip movements, muscle dynamics), but inherently compromise vividness. This creates a perceptual gap between virtual and real humans despite their local consistency. Recent advances~\citep{xu2024vasa, jiang2024loopy} map audio features and facial motion variations into shared latent spaces using diffusion models. Contemporary works~\citep{wei2025mocha, tian2025emo2, chen2025echomimic}, inspired by DiT architectures, have developed keypoint-free, text-driven end-to-end diffusion models that significantly improve expressiveness and control precision, producing lifelike facial expressions with rich emotional 
detail. OmniHuman-1~\citep{lin2025omnihuman1} introduces a unified framework integrating facial, body, and hand motions, demonstrating the potential of scaled diffusion models for comprehensive human animation synthesis. Other methods extend this paradigm by incorporating multiple speech inputs to enable multi-speaker dialogue in video generation~\citep{kong2025multitalk,huang2025bind}. However, these methods usually freeze the control conditions across the whole sequence, thereby limiting their applicability to real-time interactive synthesis.

\textbf{Real-time Video Generation.}
Real-time generation of expressive and responsive portrait videos remains a fundamental challenge. Traditional GAN-based approaches~\citep{wang2021one, qi2025chatanyone, guo2024liveportrait} employ two-stage pipelines: first extracting explicit motion representations, then rendering the final image with identity information. The emergence of diffusion models has shifted research toward implicit motion latents~\citep{xu2024vasa, drobyshev2022megaportraits, liu2024anitalker}, enabling more flexible control. During inference, lightweight generators combine predicted motion sequences with appearance features to synthesize animations at lower computational cost. However, they are constrained to fixed-length inputs, limiting the flexibility of content extrapolation. Most methods also rely on unimodal control signals and require all parameters to be specified in advance, which restricts expressiveness and real-time intervention. In addition, the lengthy DiT-based denoising process introduce significant computational burden. While the denoising procedure can be accelerated through distillation or skip paths~\citep{zhai2024motion,ma2024deepcache}, such techniques inevitably bring a trade-off between efficiency and high-fidelity synthesis. Recent innovations~\citep{low2025talkingmachines,yin2025slow,kim2025autoregressive} distill pre-trained diffusion transformers with sparse causal attention as autoregressive models for real-time performance. These advances demonstrate progression toward more efficient, high-quality, and temporally consistent video generation capabilities. To address these limitations, we propose a video generation framework based on autoregressive model, which enables real-time digital human control and synthesis under multimodal conditions.

\section{Method}
\label{sec:Methods}
Our objective is to enable real-time synthesis of expressive and controllable portrait videos under diverse multimodal conditions. The framework takes multimodal signals (e.g., audio, pose, or text) as inputs and generates coherent video sequences, ensuring low latency, spatiotemporal consistency and flexible controllability. To support this, we first construct a large-scale dialogue dataset of approximately 20,000 hours (Section~\ref{method:dataset}). We then design a highly compressive frame tokenizer and multimodal condition encoders with strong expressiveness (Section~\ref{method:representation}). Based on these components, we develop an autoregressive model for real-time rendering and interaction (Section~\ref{method:architecture}). Finally, we describe the training strategy (Section~\ref{method:tranining}) and inference strategy (Section~\ref{method:inference}) adopted in our framework.

\begin{figure*}[t] 
    \includegraphics[width=\textwidth]{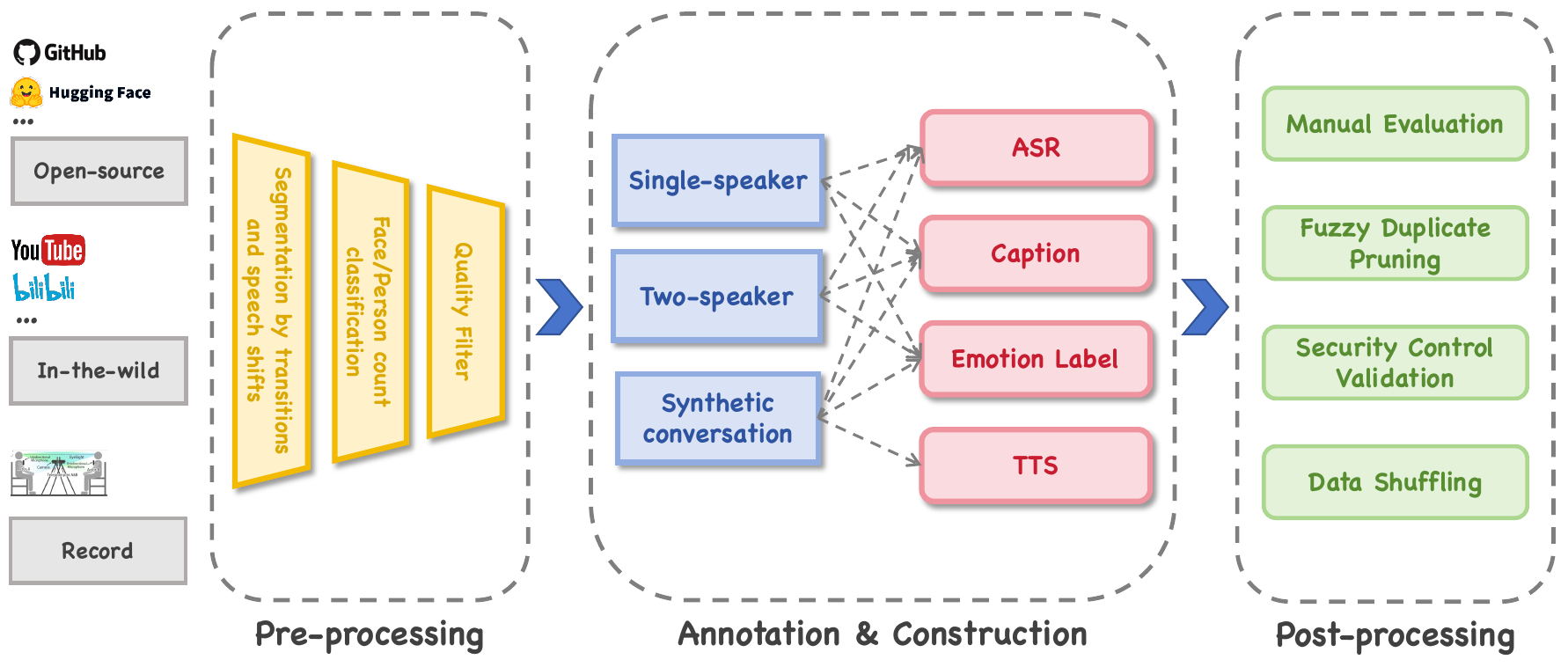}
    \vspace{-0.4cm}
    \caption{ \textbf{Overview of the dataset pipeline.} The process consists of three main stages: (1) Data collection and pre-processing; (2) Annotation and data construction; (3) Post-processing. }
    \label{fig:dataset}
\end{figure*}  

\subsection{Dataset Representation}
\label{method:dataset}
The training dataset includes single-person and two-person speech content collected from three sources: (1) publicly available benchmarks (VoxCeleb1/2~\citep{nagraniy2017voxceleb,chung2018voxceleb2}, TED-LRS~\citep{afouras2018lrs3-ted}); (2) curated online videos including podcasts, interviews, talk shows, and public speeches; and (3) custom-recorded sessions featuring controlled two-person interactions. The data processing pipeline consists of three stages: Pre-processing, Annotation and synthetic data construction, and Post-processing. The processing pipeline is illustrated in Figure~\ref{fig:dataset}.

\textbf{Pre-processing. } We apply shot boundary detection and active speaker detection (ASD) to achieve temporal segmentation, followed by filtering of human subjects via face and body detection. Each segmented clip is then subjected to rigorous evaluation in terms of visual quality, audio quality, and lip synchronization accuracy. 

\textbf{Annotation and synthetic data construction. }This stage includes quality assessment, caption generation, emotion labeling, and automatic speech recognition (ASR) transcription. A subset of single-person data is further adapted into conversational formats through semantic analysis and text-to-speech (TTS) synthesis. 

\textbf{Post-processing. }The annotated data undergoes manual review combined with automatic sampling to ensure balanced and high-quality subsets. The final dataset contains approximately 20,000 hours of pre-training video data and over 400 hours of supervised fine-tuning (SFT) data.

\begin{figure*}[t]
    \includegraphics[width=\textwidth]{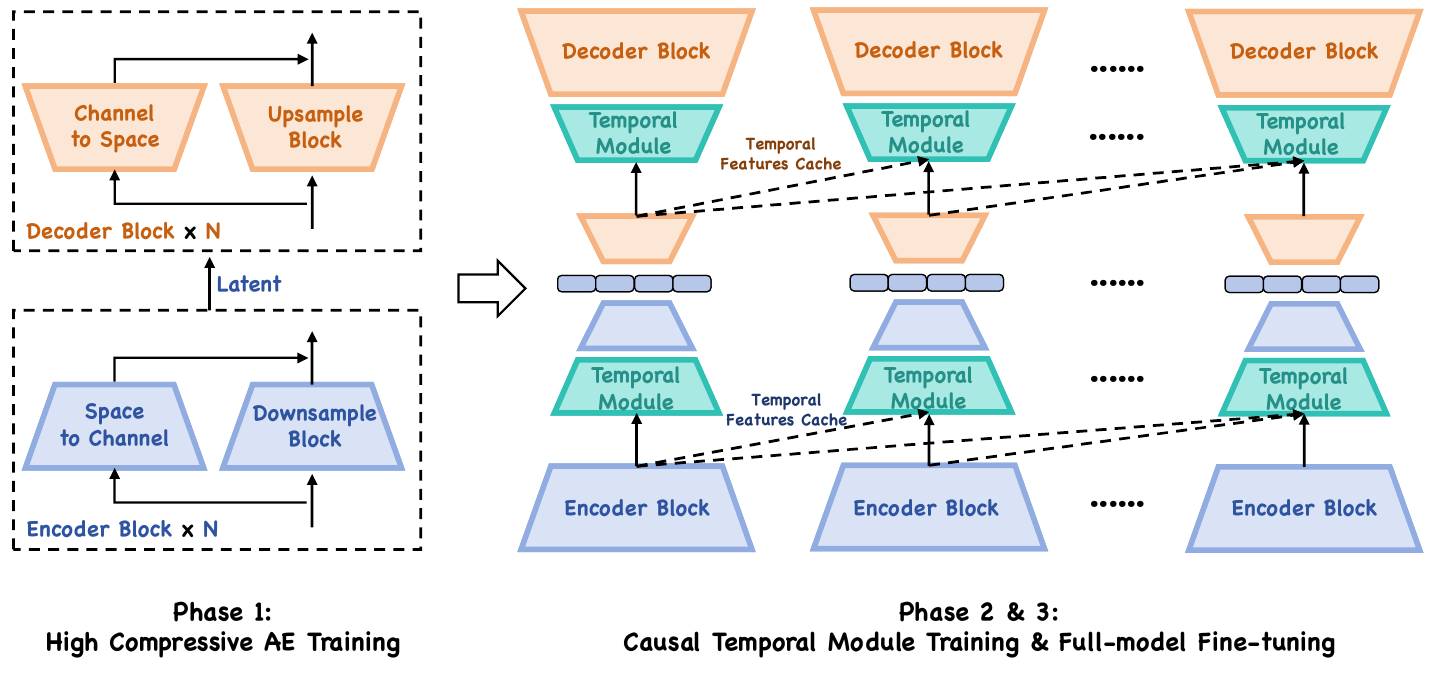}
    \vspace{-0.4cm}
    \caption{ \textbf{Illustration of our Deep Compression Autoencoder (DC-AE).} We first train the DC-AE with a spatial compression ratio of 64. In the second phase, we perform causal temporal module training. Then we apply full-model fine-tuning using an 8-frame temporal window in the third phase.}
    \label{fig:vae}
\end{figure*}  

\subsection{Multimodal Representation}
\label{method:representation}
\subsubsection{Frame Token Representation}

For real‐time AR video generation, we employ an autoencoder to transform visual inputs into compact latent representations that meet two key criteria. First, the representations must achieve satisfactory reconstruction accuracy under high spatial compression ratios, enabling efficient processing by the language model backbone. Second, to support real‐time generation, the temporal dimension remains uncompressed. However, it is crucial to model temporal dependencies during both encoding and decoding, which facilitates LLM-based autoregressive generation and ensures temporally coherent, flicker-free decoding.

To meet these requirements, we begin by training a Deep Compression Autoencoder (DC-AE) with a spatial compression ratio of 64 and 128 channels that jointly addresses spatial efficiency and temporal consistency. The model learns residuals based on the space-to-channel transformed features for efficient high spatial-compression. To capture dynamics, temporal causal 3D convolution layers and RoPE‑based attention layers are inserted after each spatial convolution layer. All temporal convolutions employ asymmetric replicate padding. In the second stage, we perform temporal module training, followed by full-model fine-tuning with an 8-frame temporal window in the third stage. During inference, we cache each frame’s temporal features (3D‑conv outputs and key/value caches) and perform streaming encoding and decoding in a frame-by-frame manner using a 5‑frame history. This streaming paradigm supports real-time, autoregressive frame-by-frame generation while preserving temporal consistency in video decoding. Although longer histories may yield better reconstructions, our 5-frame window provides a practical trade-off between computational efficiency and quality. In addition, when AR video generation is performed autoregressively with a sliding window of multiple frames, the framework can be extended to support non-causal temporal compression within each window, while maintaining causal streaming temporal encoding and decoding across windows.

\begin{figure*}[t!] 
    \includegraphics[width=\textwidth]{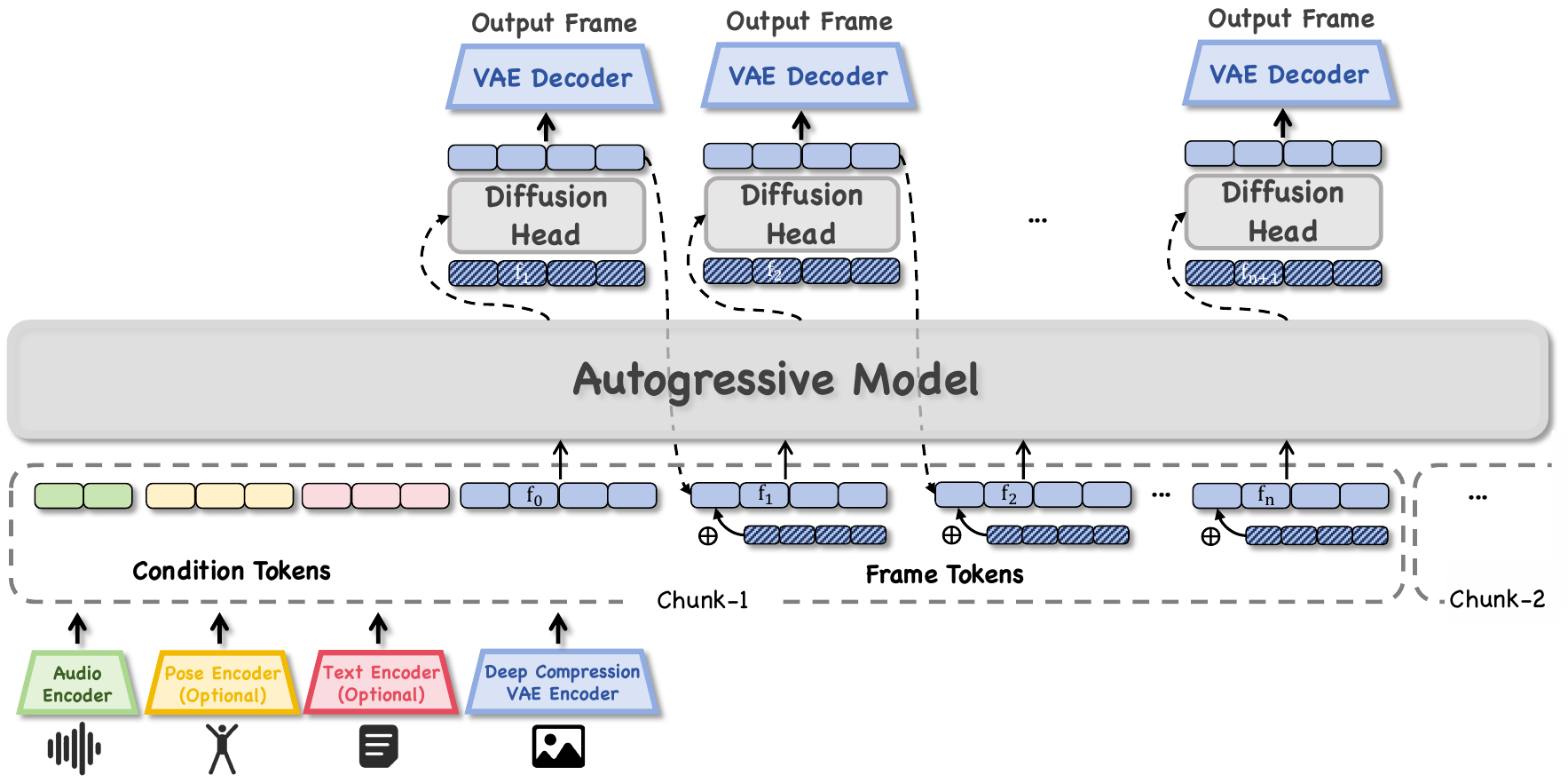}
    \vspace{-0.4cm}
    \caption{\textbf{Overview of our model architecture.} Our streaming generation framework processes inputs in chunks, where each chunk contains condition tokens (audio, pose, text) followed by frame tokens. 
    We apply teacher forcing during training, while performing next-frame prediction during inference. We corrupt frame tokens with Gaussian noise to mitigate exposure bias. The AR output serves as guidance for the diffusion head for denoising. Here, $n$ denotes the number of frames per chunk, and chunk boundaries (dashed lines) indicate positions where the inference process can be restarted with updated conditioning inputs.
    }
    \label{fig:pipeline}
\end{figure*}  

\subsubsection{Condition Representation}
Our method accepts a single portrait image and a set of control conditions spanning multiple modalities (audio, pose sequences, captions) to guide the generation process toward specific motion styles or semantic goals.

To unify these heterogeneous inputs, we introduce a \textit{multimodal condition projector} that encodes each signal into a shared latent space. For simplicity, we omit the final linear projection layer that aligns all modal representations into the same dimension.

\begin{itemize}
    \item \textbf{Audio:} $A\in \mathbb{R}^{T \times D_a} \rightarrow f_\text{audio}(A) \in \mathbb{R}^{T \times N_a \times D_h}$, where $T$ and $D_a$ denote the sequence length and the sampling rate of the audio, and $N_a,D_h$ represent the number and dimension of encoded audio tokens, respectively. We resample each 80ms audio segment to 16kHz and encode it using the Whisper-VQ~\citep{radford2022whisper} module, compressing the audio segment into a single audio token representation ($N_a=1$).
    \item \textbf{Pose:} $P \in \mathbb{R}^{T \times K_p \times D_p} \rightarrow f_\text{pose}(P) \in \mathbb{R}^{T \times N_p \times D_h}$, where $K_p, D_p$ denote the number and the dimension of the keypoints, respectively. Instead of static keypoints, we use joint velocities to represent pose trajectories. These velocities are calculated as the difference in corresponding joint positions between consecutive frames. A linear layer is employed to encode the trajectory of each frame into $N_p=10$ tokens that capture essential motion dynamics for subsequent modeling.
    \item \textbf{Text:} $X \in \mathbb{R}^{T_{ch} \times L} \rightarrow f_\text{text}(X) \in \mathbb{R}^{T_{ch} \times N_x \times D_h}$, where $T_{ch}, L$ denotes the number of chunk, the length of the textual sequence aligned with each chunk, respectively. We employ a pretrained T5 encoder~\citep{raffel2020exploring} to transform text conditions into textual embeddings with token number $N_x=256$. 
\end{itemize}

\subsection{Model Architecture}
\label{method:architecture}
\subsubsection{Autoregressive Model}
Our approach leverages an open-source large language model~\citep{qwen2.5} as the autoregressive backbone without architectural modifications, capitalizing on its inherent capabilities for sequential generation. The model's pre-trained weights and transformer architecture provide a strong foundation for modeling temporal dependencies in video sequences.

To enable efficient streaming generation, we organize inputs and outputs into logical chunks, where each chunk contains a sequence of multimodal tokens corresponding to 6 frames. Specifically, each chunk representation consists of 6 audio tokens, 60 pose tokens and 256 text tokens, concatenated sequentially from left to right. After these multimodal tokens, we append reference image tokens as well as tokens for the 6 target frames to be generated. The number of tokens per frame is determined by the spatial resolution. In our experiments, we allocate up to 60 tokens per frame, supporting image resolutions of up to $384 \times 640$. This setting can be easily scaled up to accommodate higher resolutions. Our structured token organization enables both streaming control input and sequential output generation, providing real-time responsiveness while maintaining contextual coherence across chunks, as shown in Figure~\ref{fig:pipeline}.

\begin{wrapfigure}{r}{0.5\textwidth}
    \vspace{-0.2cm}
    \includegraphics[width=0.5\textwidth]{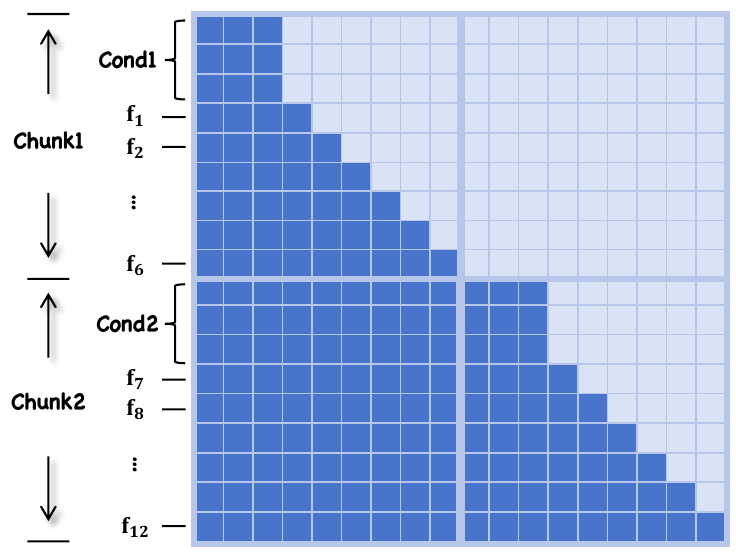}
    \vspace{-0.6cm}
    \caption{Illustration of our causal attention mask for multimodal streaming generation.}
    \label{fig:attentionmask}
\end{wrapfigure}
We also design a specialized frame-level causal attention mask for our multimodal autoregressive framework, as shown in Figure~\ref{fig:attentionmask}. Within each chunk, the condition tokens are accessible to all subsequent frame tokens, while frame tokens are restricted to attending only to the condition tokens, previous frame tokens, and their intra-frame tokens with full attention. This mask design jointly enforces temporal causality and frame-level coherence, providing a reliable mechanism for controllable multimodal generation.


For frame prediction, we employ a flow matching approach inspired by the token prediction mechanism in large language models. Our system directly predicts the representation of frame $t+1$ during the forward pass at time $t$, eliminating the need for placeholder tokens. The autoregressive generation process is formulated as
\begin{equation}
p(C, x_1, x_2, ..., x_N) = \prod_{i=1}^{N} p(x_i|C, x_1, x_2, ..., x_{i-1}),
\end{equation}

where $C$ represents the multimodal conditioning signals (audio, pose, text) and $x_i$ represents the tokens of the $i$-th video frame. This one-step prediction strategy offers three key advantages: (1) it enables straightforward teacher forcing during training, significantly improving training efficiency; (2) it allows for frame-by-frame generation without iterative refinement steps; and (3) it reduces computational overhead by avoiding multiple forward passes per frame.

\subsubsection{Diffusion Head}
Our diffusion head functions as a specialized renderer that transforms the autoregressive predictions into high-quality video frames. Drawing inspiration from MAR~\citep{li2024autoregressive}, we inject the outputs of the autoregressive model as conditioning signals into the diffusion process, but with a key distinction that we entirely eliminate mask modeling. Since the spatial relationships and semantic coherence between tokens have already been implicitly modeled by our LLM backbone, the diffusion component only needs to focus on denoising and rendering clean frames from the conditioned input.

This simplified approach enhances computational efficiency while maintaining visual fidelity. The diffusion head serves as a refinement mechanism rather than handling the full complexity of spatio-temporal modeling. For the architectural design of the diffusion component, we explored multiple options, including a standard MLP-based architecture and a more sophisticated DiT-based design~\citep{chen2023pixart}. The latter offers stronger guarantees for structural consistency in character rendering, requiring only minor modifications to the conditioning injection mechanism. 

\subsection{Training Strategy}
\label{method:tranining}
Our training approach addresses two key challenges: effective next-frame prediction and mitigation of exposure bias. We train the autoregressive backbone to predict frame $t+1$ directly from frame $t$, mapping sequences spanning frames 1 through $t$ to corresponding sequences from frames 2 through $t+1$. This forward-shifted prediction eliminates the need for placeholder tokens while maintaining temporal coherence.

Autoregressive models inherently suffer from exposure bias due to the discrepancy between training on ground-truth inputs (teacher forcing) and inference on self-generated, potentially imperfect predictions. This mismatch leads to error accumulation and rapid quality degradation over long sequences. To mitigate this gap, we follow~\citep{valevski2024diffusion} and adopt a controlled noise injection strategy during training. Context frames are corrupted by add Gaussian noise of varying magnitudes to their latents, thereby simulating the imperfect predictions encountered at inference. We simultaneously provide noise level as an conditioning signals to the model, making it adapt to varying input quality levels. 

Specifically, we uniformly set noise levels up to a maximum noise scale of 0.5, discretize them into 20 buckets, and learn a dedicated embedding for each bucket. For each batch, we randomly determine a noise level ID from 0 to 19 for each frame. These IDs are mapped to corresponding noise embeddings through a learnable embedder and expanded to match the dimensions of the frame tokens. The noise is applied by interpolating between the original latents and Gaussian noise according to the sampled corruption level, following 
\begin{equation}
x_{noisy} = \sigma \cdot \epsilon + (1 - \sigma) \cdot x,
\end{equation}
where $\sigma$ is the noise level and $\epsilon \sim \mathcal{N}(0, I)$ is the Gaussian noise. This approach systematically bridges the domain gap between training and inference by teaching the model to recover from corrupted context. It enables the network to correct errors from previously generated frames, which proves critical for maintaining frame quality and temporal consistency over long sequences.



For multimodal training, we combine diverse datasets (audio-to-video, pose-to-video, etc.) using a consistent token organization pattern: audio tokens, followed by pose tokens, text tokens, and frame tokens. For datasets missing certain conditioning signals, we employ special placeholder tokens to maintain structural consistency across heterogeneous inputs. During inference, the same token organization allows flexible combinations of different modalities as needed.

The training objective follows the flow matching formulation:

\begin{equation}
\begin{gathered}
\mathcal{L}(\theta) = \mathbb{E}_{x_0, x_1, t}[|| v_\theta(x(t), t) - v^*(x_0, x_1)||_2^2], \\
x(t)=(1-t)x_0+tx_1, \quad v^*(x_0, x_1)=x_1-x_0,
\end{gathered}
\end{equation}

where $x_0 \sim p_{data}(x), x_1 \sim p_{noise}(x)$ denote the clean target frame and noise distribution, respectively. This objective guides the model to learn the optimal vector field for transforming noisy representations to clean frames.

For long-duration generation, we introduce a specialized post-training phase using identity-specific data at higher resolution and frame rates (12.5 fps $\rightarrow$ 25fps). Unlike prior approaches that rely on vast amounts of person-specific data~\citep{ao2024body}, our method efficiently leverages limited but focused datasets to reduce error accumulation. This targeted adaptation enables stable, high-quality video synthesis over extended periods (up to one hour), with minimal quality degradation compared to pretraining-only baselines.

\subsection{Inference Strategy}
\label{method:inference}
During inference, we organize our generation process into temporal chunks of 480ms to enable efficient streaming. Within each chunk, the autoregressive model sequentially generates frame tokens, which are then passed to the diffusion head for denoising. After completing one chunk, the model proceeds to the next, enabling continuous video generation without the need of processing the entire sequence at once.

To ensure visual consistency, we apply a uniform noise level to all frames within each inference step, which can be adjusted to optimize quality. Even with minimal added noise, we observe significant improvements in temporal stability compared to standard autoregressive generation. This controlled noise design ensures temporal coherence while simplifying the denoising process.

For extended inference periods, we implement a frame truncation strategy inspired by FAR~\citep{gu2025far}, which recognizes that historical frames beyond a certain temporal distance cease to provide useful information and may instead introduce error accumulation. By limiting the context window to the most recent 8 chunks and discarding low-relevance historical frames, our system effectively reduces cumulative errors while preserving essential temporal dependencies for coherent motion synthesis.

\section{Experiments}
\label{sec:Exp}
\begin{figure*}[t] %
    \includegraphics[width=\textwidth]{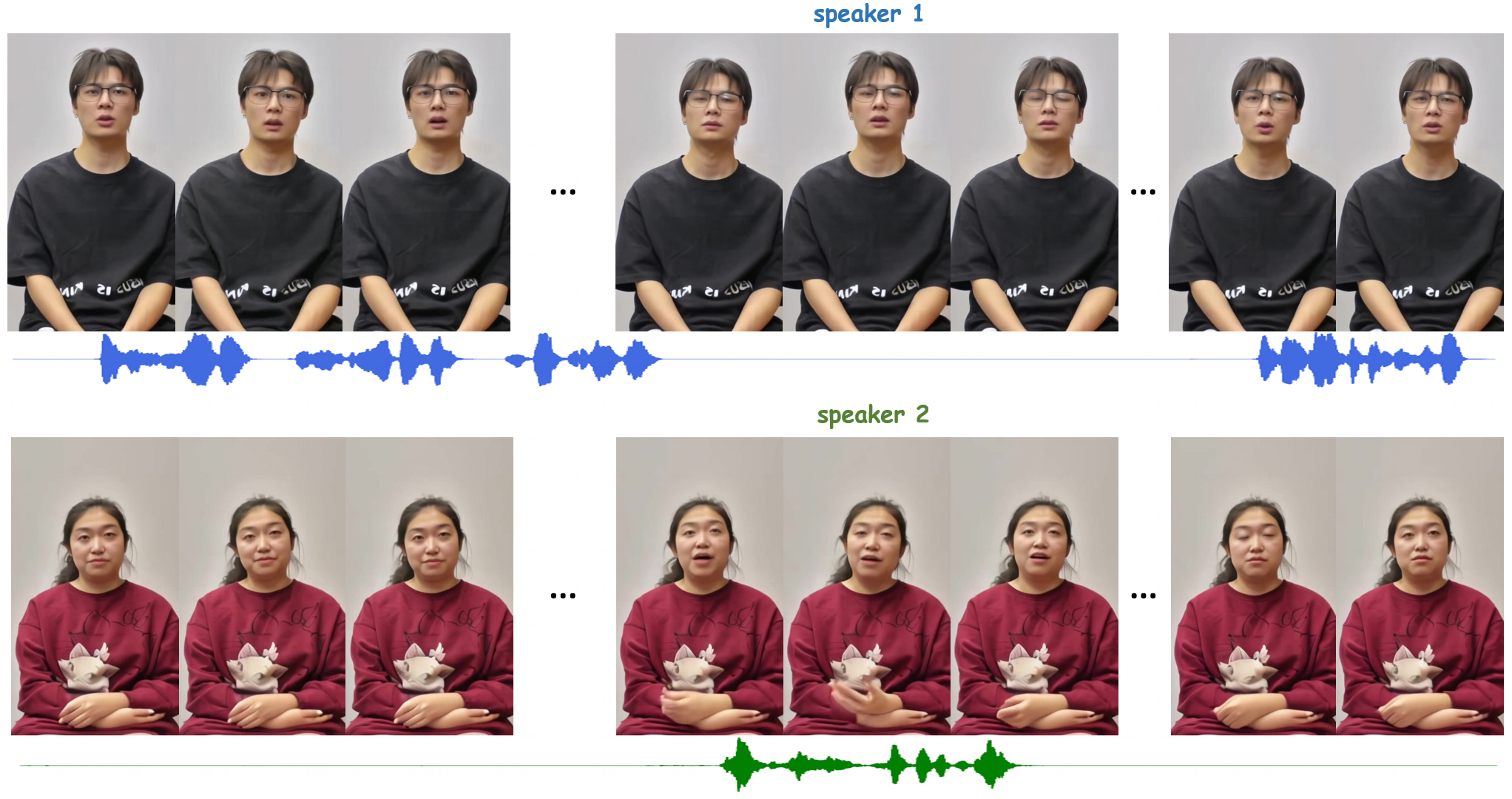}
    \vspace{-0.4cm}
    \caption{
    \textbf{Multi-speaker digital conversation with audio-driven avatars.} Speaker 1 (top) and Speaker 2 (bottom) demonstrating turn-taking dialogue, with corresponding audio waveforms shown in blue and green.
    }
    \label{fig:dialog}
\end{figure*} 

\begin{figure*}[t] %
    \includegraphics[width=\textwidth]{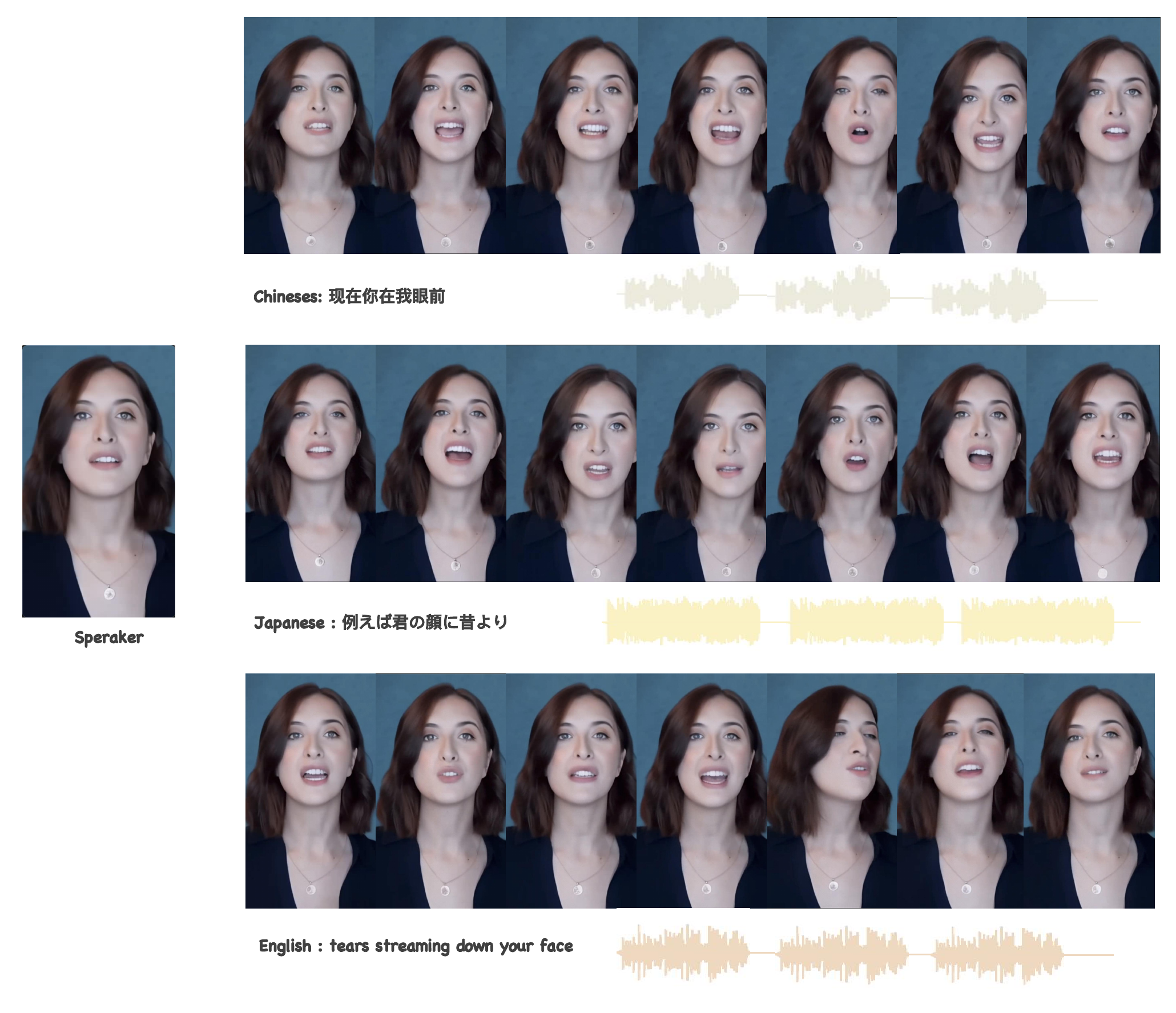}
    \vspace{-0.4cm}
    \caption{\textbf{Cross-lingual singing generation with synchronized lip movements.} Our model accurately renders lip synchronization across multiple languages, demonstrating the model's fine-grained understanding of cross-lingual phoneme-to-viseme mapping without explicit language identification.}
    \label{fig:multi-language}
\end{figure*} 

\begin{figure*}[t] %
    \includegraphics[width=\textwidth]{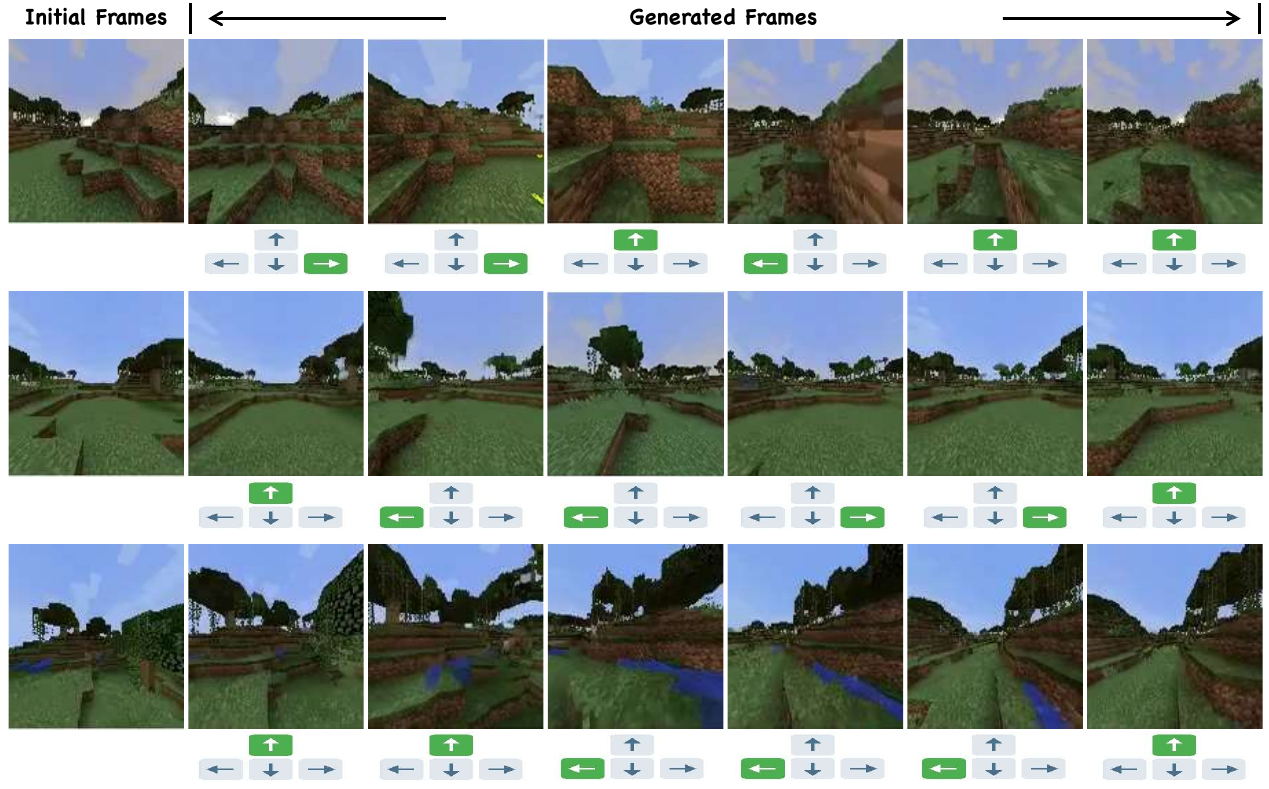}
    \vspace{-0.4cm}
    \caption{\textbf{General controllable video generation on Minecraft dataset.} By incorporating directional control into our multimodal condition encoding, we realize a real-time interactive world model that exhibits remarkable visual consistency and memory capabilities.}
    \label{fig:minecraft}
\end{figure*} 

\subsection{Implementation details.}
Our model is trained using the DeepSpeed ZeRO-2 optimization framework across 64 NVIDIA H800 GPUs for approximately 7 days. We adopt bfloat16 precision and apply gradient clipping at 1.0 to ensure numerical stability. We use AdamW as the optimizer with a learning rate of $5.0\times10^{-5}$ for both the main model and pretraining components, combined with a cosine learning rate scheduler, 1000 warmup steps, and weight decay of 0.01. The training is performed for a total of 100 epochs with a per-device batch size of 2. The autoregressive backbone is based on Qwen2.5-3B~\citep{qwen2.5}, while the diffusion head follows PixArt-$\alpha$ architecture~\citep{chen2023pixart} with approximately 0.5B parameters. We adopt 4 denoising steps during both training and inference, which we find to provide an effective balance between efficiency and synthesis quality.



\subsection{Qualitative Results.}

In the experiments, we present real-time digital human generation results driven by audio and the reference image. We leave pose and text conditioning as future work, which can be readily scaled up by incorporating corresponding modality encodings into our framework.

\textbf{Duplex conversation.} Figure~\ref{fig:dialog} showcases our system enabling natural turn-taking dialogue between digital avatars with synchronized audio-visual responses. Each avatar displays appropriate listening behaviors when the other is speaking, and becomes animated with synchronized lip movements and facial expressions when driven by its corresponding audio input. The audio waveforms (visualized in blue and green) clearly delineate the speaking turns. This demonstrates our model's abilities to generate contextually appropriate reactions and maintain speaker identity while handling the complex dynamics of conversational interaction.

\textbf{Multi-language Long Video Synthesis.} We present an example of cross-lingual singing synthesis in Figure~\ref{fig:multi-language}, where we achieve precise phoneme-level synchronization across diverse language families. Our system can generate convincing lip movements for songs in Mandarin Chinese, Japanese, and English, adapting to the distinct phonetic characteristics of each language without requiring language-specific training. Notably, our model is capable of generating videos up to 4 minutes in length without significant drift. Please refer to our project page for video visualizations.

\textbf{General Interactive Video Generation.} Our model architecture can also flexibly accomodate arbitrary modal conditions as inputs, making it seamlessly applicable to general interactive video generation tasks. By reformulating multimodal conditions into directional control signals and training on the Minecraft dataset~\citep{yan2023temporally}, our approach effectively serves as a real-time interactive world model. As shown in Figure~\ref{fig:minecraft}, our world model achieves strong visual consistency and exhibits notable memory capabilities.

\section{Conclusion}
\label{sec:Conclusion}
In this work, we present MIDAS, a multimodal interactive digital human synthesis framework for real-time video generation. Our approach builds on an LLM-based autoregressive model combined with a lightweight diffusion head, enabling low-latency, streaming synthesis under diverse multimodal controls. We additionally introduce a controlled noise injection strategy to mitigate the exposure bias between training and inference. To support training, we construct a large-scale dialogue dataset from multiple sources, and design a deep compression autoencoder to ease the long-horizon inference burden. Extensive experiments on duplex conversation, multilingual human synthesis and general interactive world model validate the effectiveness of our framework, demonstrating its advantages in responsiveness, controllability, and temporal stability. We believe that our exploration provides a solid step toward scalable and interactive digital human generation and will inspire future research in this emerging area.


\clearpage
\bibliography{iclr2025_conference}
\bibliographystyle{iclr2025_conference}


\end{document}